\documentclass[journal]{IEEEtran}
\usepackage{amssymb,amsmath}
\usepackage{color}
\usepackage{graphicx}
\usepackage{subcaption}

\usepackage{soul}  
\usepackage{ulem} 

\usepackage{url}
\usepackage{hyperref}

\usepackage{multirow}
\usepackage{multicol}
\usepackage{subcaption}

\usepackage{algorithm}
\usepackage{algorithmic}
\usepackage{booktabs}
\usepackage{times}
\usepackage{enumitem}

\usepackage{graphicx}  
\usepackage{array}     

\makeatletter

\newcommand{\Rmnum}[1]{\expandafter\@slowromancap\romannumeral #1@}
\makeatother

\newcommand{\Abf}{\pmb{A}}\newcommand{\Cbf}{\pmb{C}}\newcommand{\Ebf}{\pmb{E}}\newcommand{\Hbf}{\pmb{H}}\newcommand{\Rbf}{\pmb{R}}
\newcommand{\abf}{\pmb{a}}\newcommand{\ebf}{\pmb{e}}\newcommand{\hbf}{\pmb{h}}\newcommand{\lbf}{\pmb{l}}\newcommand{\nbf}{\pmb{n}}\newcommand{\pbf}{\pmb{p}}\newcommand{\rbf}{\pmb{r}}\newcommand{\ubf}{\pmb{u}}\newcommand{\vbf}{\pmb{v}}\newcommand{\xbf}{\pmb{x}}

\begin{document}


\title{Radar and Event Camera Fusion for Agile Robot Ego-Motion Estimation}

\author{Yang~Lyu,~\IEEEmembership{Member,~IEEE}, Zhenghao Zou, Yanfeng Li, Xiaohu Guo, Chunhui Zhao, Quan Pan,~\IEEEmembership{Member,~IEEE}
\thanks{Yang Lyu, Zhenghao Zou, Yanfeng Li, Xiaohu Guo, Chunhui Zhao, Quan Pan are with the School of Automation, Northwestern Polytechnical University, Xi'an,
Shaanxi, 710129 P.R. China. e-mail: lyu.yang@nwpu.edu.cn.}

\thanks{This work was supported
by the National Natural Science Foundation of China under Grant 62203358, Grant 62233014, and Grant 62073264. (\it {Corresponding author: Yang Lyu})

} 
}

\maketitle

\begin{abstract}
Achieving reliable ego motion estimation for agile robots, e.g., aerobatic aircraft, remains challenging because most robot sensors fail to respond timely and clearly to highly dynamic robot motions, often resulting in measurement blurring, distortion, and delays. In this paper, we propose an IMU-free and feature-association-free framework to achieve aggressive ego-motion velocity estimation of a robot platform in highly dynamic scenarios by combining two types of exteroceptive sensors, an event camera and a millimeter wave radar,  First, we used instantaneous raw events and Doppler measurements to derive rotational and translational velocities directly. Without a sophisticated association process between measurement frames, the proposed method is more robust in texture-less and structureless environments and is more computationally efficient for edge computing devices. 
Then, in the back-end, we propose a continuous-time state-space model to fuse the hybrid time-based and event-based measurements to estimate the ego-motion velocity in a fixed-lagged smoother fashion. 
In the end, we validate our velometer framework extensively in self-collected experiment datasets featured by aggressive motion and HDR light conditions. The results indicate that our IMU-free and association-free ego motion estimation framework can achieve reliable and efficient velocity output in challenging environments.
The source code, illustrative video and dataset are available at \url{https://github.com/ZzhYgwh/TwistEstimator}.
\end{abstract}


\IEEEpeerreviewmaketitle

\section{Introduction}
Reliable ego-motion estimation is fundamental to autonomous robotic platforms. Early solutions rely on GNSS/INS, while more recent SLAM-based methods integrate diverse sensors such as cameras, LiDARs, and radars, making them more adaptable and widely applicable.
Successful deployments of SLAM-based approaches on various platforms utilize combinations of sensors such as cameras, LiDARs and IMUs, leveraging their measurements to fully resolve all degrees of freedom in platform's pose estimation.
Nevertheless, highly agile robotic systems, such as aerobatic UAVs, racing UGVs, and jointed robots demand fast and accurate velocity-state feedback to maintain stability, enable aggressive motion control \cite{fu2022coupling}, and support timely decision-making under high-dynamic conditions \cite{ji2022concurrent}. Real-time velocity feedback can significantly reduce state update latency, for example, within a UAV flight control stack\cite{cao2025proximal} or motion planning \cite{fan2019mid}. Conventional perception sensors typically fail to capture instantaneous velocity, and odom-based or SLAM-based methods suffer from substantial latency \cite{pang2024radarmoseve}. Even when combined with high-frequency inertial data, these approaches cannot ensure an immediate response to rapid state changes. Therefore, real-time velocity estimation is crucial for robust closed-loop control and reactive, decision-oriented autonomy in highly dynamic robotic systems.



\begin{figure}[t]
    \centering
    \includegraphics[width=1.0\linewidth]{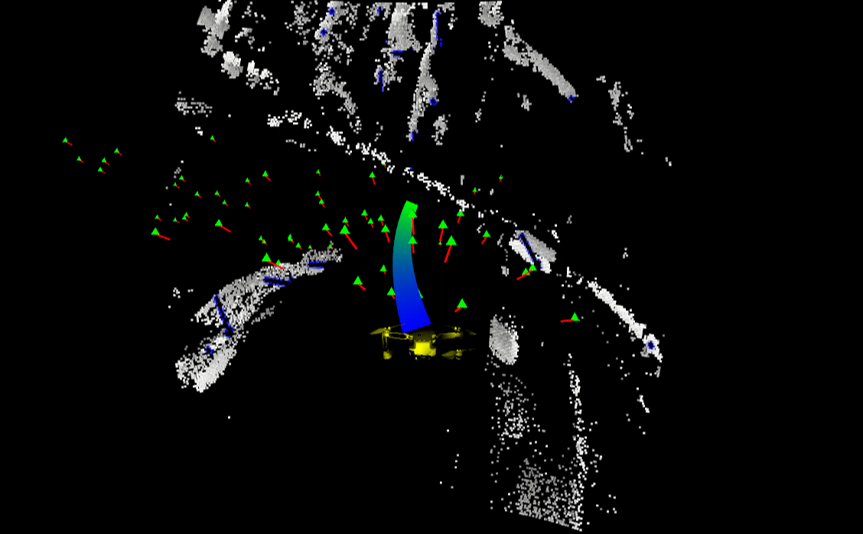}
    \caption{Robot Ego-Motion Estimation. The proposed pipeline estimates the 6-DoF instantaneous velocity of a rapidly moving robot by fusing multi-point MMWave Radar Doppler measurements—including anchors (Green) and Doppler (Red) with sparse event-based normal optical flow (Blue) computed on the SAE (White). By integrating the estimated velocities, the 0.5-second motion trajectory is reconstructed as a Color-Gradient Line(Blue for near, Green for far). }
    \label{fig:Task}
\end{figure}



Cameras capture images over an exposure period, and most LiDARs perform ranging sequentially over time. For platforms with high translational and rotational velocities, motion during sensing cannot be neglected, as it causes blur and distortion in the measurements. In contrast, recently developed event cameras asynchronously output pixel measurements by generating events in response to changes in light intensity, providing motion-sensitive signals for ego-motion estimation. Moreover, event measurements inherently encode velocity information in the image domain, establishing a direct and natural link to instantaneous velocity estimation\cite{lu2025event}. However, event cameras alone lack metric scale and cannot fully recover the 6-DoF motion of a mobile platform.

\begin{figure*}[t]
    \centering
    \includegraphics[width=1.0\linewidth]{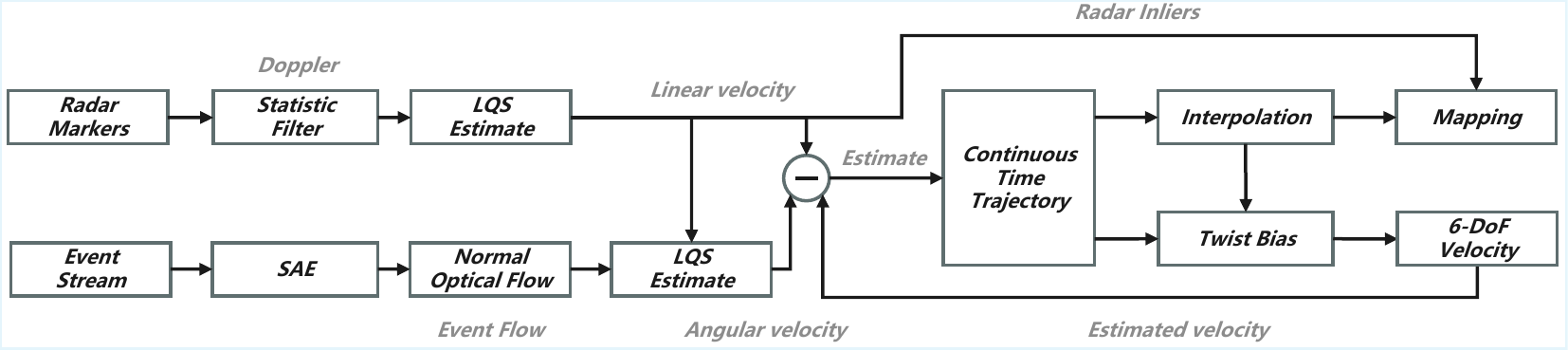} 
    \caption{The proposed ego-motion estimation pipeline.} 
    \label{fig:pipeline}
    \vspace{-0.4 cm}
\end{figure*}

In this paper, we aim to develop a framework that combines an event camera with a complementary 4D millimeter wave radar, rather than an IMU, to estimate the ego-motion velocity of an agile moving robot, as illustrated in Fig. \ref{fig:Task}.
Specifically, we use the instantaneous Doppler measurement to provide the metric information. A detailed pipeline of our proposed method is presented in Fig. \ref{fig:pipeline}.
To our knowledge, this work is the first ego-motion estimation attempt based on such a sensor setup.
By leveraging the complementary and instantaneous measurement characteristics of the sensor setup, this framework avoids the need for computationally expensive frame-to-frame feature matching and therefore improve the efficiency and robustness in challenging sensing environments. Moreover, the velocity from radar Doppler is considered drift-free compared to that from IMU measurement, and therefore can provide more accurate velocity, even during long-term estimation.

A key consideration in our sensor setup is the combination of low-rate radar Doppler measurements with high-rate event-based camera data. While radar Doppler signals are nominally low-frequency, in practice, their incremental constraints on translational velocity are sufficient, since translational motion typically evolves more smoothly than rotational motion in most agile scenarios. High-frequency event-based measurements provide dense temporal constraints on motion, particularly for angular changes, which complements the radar’s contributions. Moreover, our continuous-time state parameterization inherently accommodates asynchronous measurement streams without requiring explicit time alignment. This design mitigates potential latency or computational imbalance issues associated with naive discrete-time fusion, and ensures that the hybrid sensor setup delivers accurate 6-DOF velocity estimates at high frequency. Consequently, explicit synchronization of radar and event measurements is unnecessary, and the proposed method maintains reliable real-time performance under high-dynamic conditions.
The contributions are as follows:
\begin{itemize}
 \item 
We develop a lightweight 3D ego-motion estimation front-end that directly
derives instantaneous and drift-free metric linear and angular velocities from event-based pixel-level dynamics and the Doppler effect of radar point clouds, and avoids frame-to-frame associations. These associations are highly susceptible to failure in
scenarios where texture and structural features are insignificant, as well as
during high-maneuver movements.
 
\item We develop a continuous-time back-end to support the fusion of the high-
rate asynchronous event-based measurement and the low-rate periodic
radar measurements, as well as other potential ego-motion measurements
(e.g. IMU measurements), to generate high-speed motion estimation that matches high-speed
agile maneuvers.
\item We evaluate our proposed method on various self-collected data sequences, including both aggressive motions and HDR situations, and compare it against several state-of-the-art approaches, further demonstrating the advantages of our sensor setup and method.
\end{itemize}

The remainder of the paper is organized as follows. Section II provides a review of related literature. The association-free and IMU-free velocity estimation front-end is derived in Section III. Section IV formulates the continuous-time ego-motion estimator. Validations of the proposed framework are provided in Section V. Section VI concludes the paper.
\section{Related Works}
\subsection{Event-camera based ego-motion estimation}

Event cameras provide high temporal resolution, wide dynamic range, and low power use, making them well-suited for vSLAM in challenging settings. 


Event-based odometry methods can be broadly categorized by how event data is processed. Some works, such as \cite{rebecq2017real} and \cite{zihao2017event}, convert asynchronous events into event frames within spatio-temporal windows for feature detection and tracking. To enhance geometric constraints, line features are further explored in \cite{chamorro2020high,chamorro2022event}, while \cite{xu2023tight} integrates line features with IMU data to achieve high-frequency motion estimation.

Some works aim to directly estimate motion from events, bypassing traditional feature detection and tracking. A simple approach generates event frames through spatio-temporal alignment \cite{rebecq2016evo}. More advanced methods, like \cite{hidalgo2022event}, tightly fuse events and frames using EGM and PBA. 

With deep learning’s rise in computer vision, applying it to event camera SLAM is promising. Event data is usually converted into frame-like forms for CNN processing.
\cite{zhu2019unsupervised} introduces an unsupervised method that encodes events as temporal volumes and estimates optical flow, ego-motion, and depth from motion blur. In a supervised approach, \cite{gehrig2021combining} presents a network that fuses asynchronous events and monocular images for continuous-time depth and motion estimation.
DEVO\cite{klenk2023devo} advances event-based odometry but depends on the quality of trained event patches for accurate motion estimation.


Building on multi-sensor fusion, \cite{zhu2023event} combines event and RGB-D data to improve pose estimation for agile-legged robots under dynamic maneuvers, and \cite{wang2024fast} demonstrates that tightly coupling LiDAR, IMU, event, and standard camera measurements enables real-time and robust odometry in challenging conditions. These methods generally leverage the dynamic nature of event data to enhance tracking robustness and accuracy. PL-EVIO\cite{guan2023pl} achieves robust, real-time state estimation by tightly fusing asynchronous event streams, standard images, and inertial measurements.

Although these methods show strong performance, their reliance on frame-to-frame association can be computationally demanding for onboard systems with limited resources. In contrast, due to its sensing mechanism, an event camera provides instantaneous motion cues. Thus, we explore deriving instantaneous velocity directly from event data.
\vspace{-0.3cm}
\subsection{Radar ego-motion estimation}
A mmwave radar can provide two types of information to achieve ego-motion estimation. 


First, relative transformations can be obtained by registering radar point clouds between frames or with a map. A radar SLAM system in \cite{hong2022radarslam} shows strong performance in all-weather conditions. Similarly, \cite{barnes2020under} proposes an unsupervised method for feature detection and tracking, followed by odometry estimation. \cite{DoerENC2020} introduces a radar-inertial odometry (RIO) system with online extrinsic calibration for robust localization. \cite{burnett2024continuous} presents a continuous-time radar-inertial framework that fuses IMU data with spinning radar and models radar point uncertainty to improve real-time accuracy and robustness, and \cite{kim2024radar4motion} proposes a 4D radar odometry method leveraging Doppler and RCS with weighted scan-to-submap matching for IMU-free ego-motion estimation. However, these point cloud registration front-ends often incur high computational costs, posing challenges for resource-constrained robotic platforms.

Radar uniquely offers instantaneous radial velocity through the Doppler effect, enabling direct ego-motion estimation. 
Compared to the IMU, the velocity from the Doppler effect can be treated as a drift-free measurement, even in long-term sensing environments.
Studies\cite{kellner2013instantaneous, kellner2014instantaneous} estimate vehicle velocity from one or multiple radars, while \cite{park20213d} incorporates a 3D velocity factor in pose graph optimization.
These methods reduce computation by avoiding point cloud pre-processing but can face error accumulation during pose estitmation and have limited angular velocity observability.


Recent radar-based SLAM frameworks combine point and velocity information for robust pose estimation. \cite{zhuang20234d} formulates frame-to-map registration and Doppler velocity as pose and velocity prior factors. Similar approaches are adopted in \cite{zhang20234dradarslam} and \cite{li20234d}, using different velocity integration strategies. A related framework, \cite{nissov2024roamer}, fuses Doppler LiDAR, IMU, and velocity in a graph-based optimization.

In this paper, we aim to achieve ego-motion with limited computation and storage resources by leveraging Doppler velocity. Specifically, we consider removing the angular unobservability by combining an event camera. With the two instantaneous measurements, registration-free 6-DOF velocities can be recovered directly without the aid of an inertial measurement unit (IMU).
\vspace{-0.3cm}
\subsection{Continous-time Representation}

Continuous-time (CT) based localization has gained popularity, especially with multi-sensor fusion becoming standard. Furgale \textit{et al.}\cite{furgale2012continuous} first introduced representing trajectories as Gaussian bases in a CT SLAM framework. Subsequent works extended this to various sensor setups. For example, \cite{hug2022continuous} presents a CT SLAM with asynchronous stereo-inertial sensors, avoiding IMU pre-integration via spline-based trajectories. A LiDAR-only odometry using CT formulation is proposed in \cite{zheng2024traj}, enabling pose estimation during aggressive motions by processing high-frequency streaming LiDAR points without motion compensation.
Closely related to our work, \cite{mueggler2018continuous} fuses asynchronous event camera data with IMU in a CT SLAM, effectively handling high-rate measurements for dynamic platforms. Additionally, CT SLAM frameworks like \cite{lv2023continuous} support multi-sensor fusion (LiDAR, camera, IMU) with online time-offset estimation, demonstrating CT’s flexibility in handling complex sensor fusion tasks.

In this paper, we plan to achieve ego-motion estimation based on the fusion of an event camera and a Doppler radar, which outputs high-frequency 6-DOF velocities for an agile robot. However, the sampling rate of the radar Doppler measurement is much lower than the events, and may not match the agility of a robot's motion pattern. 
Thanks to the nature of handling asynchronous measurement of the CT-SLAM, we can still fuse them in a CT fashion, and the status parameterization method allows us to output high-frequency estimates.

\vspace{-0.4cm}
\subsection{Notations}
In this paper, we use lowercase and uppercase bold letters to represent vectors and matrices, respectively. Time in continuous- and discrete-time is denoted by $t\in \mathbb{R}^+\cup \{0\}$ and $k\in \mathbb{Z}$. Specifically, we use $(\cdot)(t)$ and $(\cdot)^{(k)}$ to represent variables in the continuous-time domain and the discrete time, respectively. In
addition, we use calligraphic font letters to denote variables in different frames. We use $\mathcal{G}$ to represent the global frame, and $\mathcal{R}$ and $\mathcal{E}$ are the radar frame and camera frame respectively. For example, $^{\mathcal{G}}\pbf_r(t)$ denotes the position of the radar in the global frame at time $t$, and $^\mathcal{R}\pmb{\omega}_r^k$ is the radar's rotation velocity at time instance $k$.

\section{Velocity Front-ends}
\begin{figure}[!ht]
    \centering
    \includegraphics[width=1.0\linewidth]{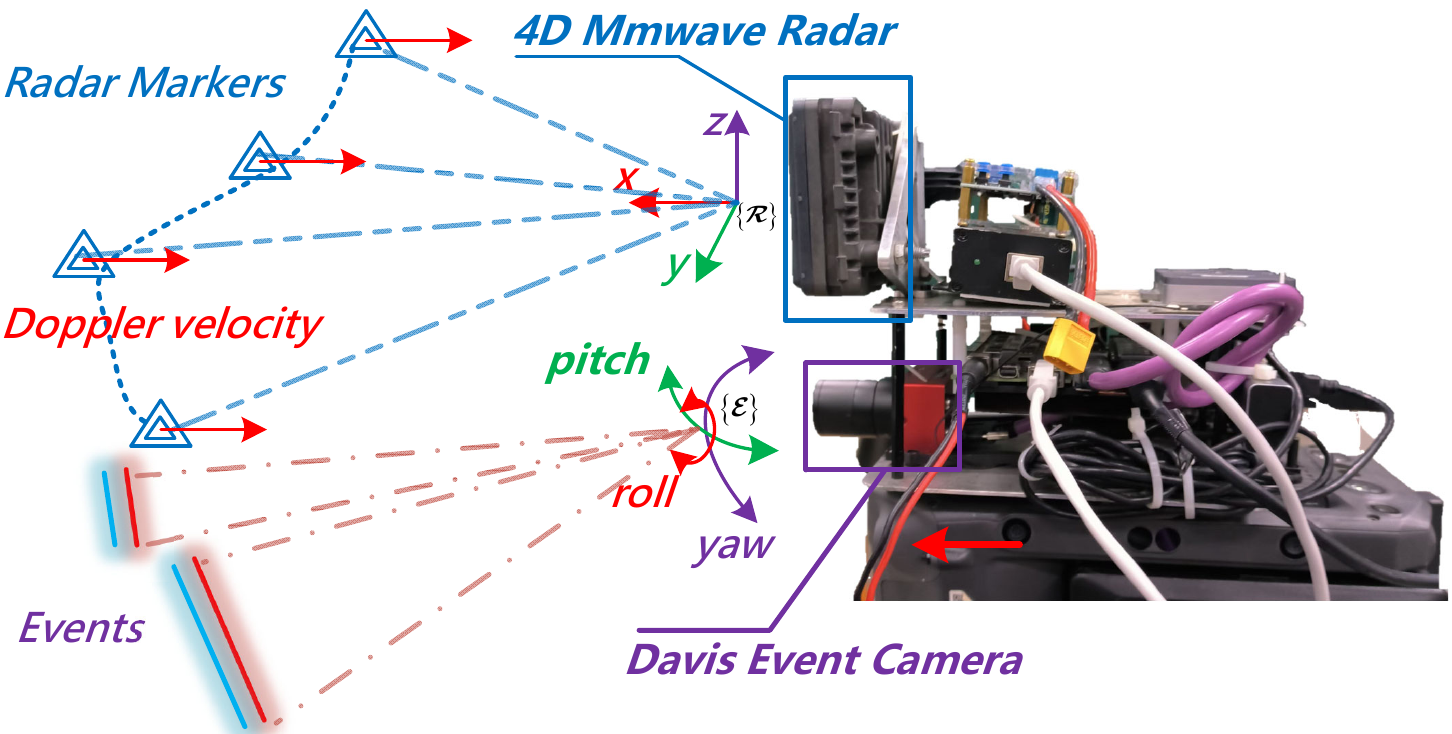} 
    \caption{Coordinate Systems.} 
    \label{fig:coordinate}
    \vspace{-0.6 cm}
\end{figure}

In this section, the ego-motion front end is provided. Specifically, we derive linear and angular velocity from a 4D MMWR and an event camera, respectively. The coordinate system and system setup are illustrated in Fig.\ref{fig:coordinate}.


\subsection{Linear Velocity from 4D Radar}

The 3D ego-motion velocity of a 4D millimeter-wave radar in its local frame can be obtained based on the position of the point clouds and their Doppler velocities.

Given one radar point cloud frame with point set as $C$, the coordinate of a static 3D world point $i\in C$ in the radar local frame $\{\mathcal{R}\}$ is defined as $^{\mathcal{R}}\pmb{p}_{i}\in \mathbb{R}^3$, the corresponding Doppler velocity measurement is defined as
\begin{equation}
	v_{i}=-\frac{^{\mathcal{R}}\pbf_{i}^\top}{\|^{\mathcal{R}}\pbf_{i}\|} {}^{\mathcal{R}}\vbf_{\text{r}}, 
	\label{velcotyu2}
\end{equation}
\label{velocity_sec}
where ${}^{\mathcal{R}}\vbf_{\text{r}}$ is the ego-motion velocity of the radar in its local frame. 
Given $n\ge 3$ valid points, the ego-motion velocity ${}^{\mathcal{R}}\vbf_{\text{r}}$ can be estimated in a least-square manner with
\begin{equation}
 \Hbf^{\mathcal{R}}\vbf_{{r}} =\vbf_{d},
\end{equation}
where $\Hbf\triangleq \begin{bmatrix}
\frac{{}^{\mathcal{R}} \pbf_1}{\|{}^{\mathcal{R}}\pbf_1\|} &\frac{{}^{\mathcal{R}}\pbf_2}{\|{}^{\mathcal{R}}\pbf_2\|} & \cdots & \frac{{}^{\mathcal{R}}\pbf_n}{\|{}^{\mathcal{R}}\pbf_n\|}
\end{bmatrix}^\top$, and $\vbf_{d} \triangleq \begin{bmatrix}
		-v_1&-v_2&\cdots& -v_n 
	\end{bmatrix}^\top$ is the stacked Doppler velocities of all points in a frame. 

Practically, the velocity accuracy of the above estimation is often degraded due to outliers from moving objects or clutter. To mitigate the effects of Doppler measurement noise, we apply RANSAC-based outlier rejection. Furthermore, we need to evaluate the uncertainties of each dimension based on the distribution of valid points in each frame, as below:
\begin{equation}
 \pmb{\Sigma}_{\vbf}={\sigma}^2\left(\Hbf^{\top} \Hbf\right)^{-1}+\left(\Hbf^{\top} \Hbf\right)^{-1}\left(\sum_i J_i^{\top} \Sigma_{p, i} J_i\right)\left(\Hbf^{\top} \Hbf\right)^{-1},
\end{equation}
where $J_i = {\partial \hbf_i} / {\partial \pbf_i}$, and $\hbf_i = - {\pbf_i} / {\|\pbf_i\|}$. $\Sigma_{p, i}$ is the position covariance of each point calculated from the noise variances of azimuth angle, elevation angle, and range measurements of each point $i$. 

\subsection{Angular Velocity from Event Camera}
In this section, we use instantaneous event camera data to derive the angular velocity. First, we obtain normal flow directly from raw events, and then we derive the angular velocity based on a continuous-time epipolar constraint. 

\subsubsection{Surface of Active Events (SAE)}

Event cameras produce asynchronous events $\ebf_i = (x_i, y_i, t_i, p_i)$ when pixel brightness changes, where $(x_i, y_i)$ is the pixel location, $t_i$ is the timestamp, and $p_i \in \{-1,+1\}$ is the polarity. To compactly represent recent activity, we construct a \textit{Surface of Active Events} (SAE), a 2D map ${T} \in \mathbb{R}^{H \times W}$, where each pixel stores the latest event timestamp:

\begin{equation}
    {T}(x, y) = \max_{i \in \mathcal{E}_{x,y}} t_i,
\end{equation}
with $\mathcal{\Ebf}_{x,y} = \{ i \mid (x_i, y_i) = (x, y) \}$ the set of events at pixel $(x, y)$. Each incoming event $\mathbf{e}_i$ updates the SAE at its location:
\begin{equation}
    {T}(x_i, y_i) \leftarrow t_i,
\end{equation}

providing a continuously updated temporal surface for motion estimation. To approximate the true velocity, SAE is constructed using events within a short temporal window of 30–50 ms.

\subsubsection{ Normal Flow on SAE}
According to \cite{lu2023event}, only normal flow, which is defined as the part of optical flow parallel to the image gradient, can be recovered from the SAE. To obtain the full optical flow, we assume a patch of pixel area in SAE shares a common optical flow, and there are multiple edges that can be detected within a short time period.
According to the last subsection for SAE, $ {T}(\ubf):\mathbb{R}^2\to \mathbb{R}$ where $\ubf\in \mathbb{R}^2$ denotes the position of the pixel on the image plane. The function returns the time stamp of the latest report event of a specific pixel $\ubf$. When an edge moves with a constant normal velocity $v_n\in \mathbb{R}^+$, a simple model of ${T}(\ubf)$ can be defined as 
\begin{equation}
     {T}(\ubf)=t-\frac{d(\ubf)}{v_n},
\end{equation}
where $t$ is the current time, and $d(\ubf)= \ubf^\top\nbf$ is the signed distance of a point $\ubf$ from the edge. $\pmb{n}$ is a unit vector normal to the edge and pointing in the direction of increasing time.

Taking the gradient of $T(\ubf)$, we have 
\begin{equation}\label{eq:gradient}
    \nabla T(\ubf)=-\frac{1}{v_n} \nabla d(\ubf) =-\frac{\nbf}{v_n}.
\end{equation}
Then taking the norm of both sides, 
\begin{equation}
\|\nabla T\|=\frac{1}{v_n} \quad \Longrightarrow \quad\|\dot\ubf_n\|= v_n=\frac{1}{\|\nabla T\|}.
\end{equation}

Given the raw events, $\nabla T$ can be obtained by fitting a local spatio-temporal plane on SAE and then calculating its gradient \cite{valeiras2018event}. Based on the definition of normal flow, we have 
\begin{equation}
    \frac{\nabla T_1}{\|\nabla T_1\|}\dot\ubf = \frac{1}{\|\nabla T\|}.
\end{equation}

Given more than two edges within a patch, a local optical flow can be obtained in a least-square fashion based on the following equations
\begin{equation}
    \begin{bmatrix}
         \frac{\nabla T_1}{\|\nabla T_1\|}\\
         \frac{\nabla T_2}{\|\nabla T_2\|}\\
        \vdots\\
         \frac{\nabla T_m}{\|\nabla T_m\|} 
    \end{bmatrix}\dot{\ubf} = \begin{bmatrix}
        \frac{1}{\|\nabla T_1\|}\\ \frac{1}{\|\nabla T_2\|}\\\vdots\\
         \frac{1}{\|\nabla T_m\|}
    \end{bmatrix}.
\end{equation}

\subsubsection{Angular Velocity from Normal Flow}
The linear and angular velocity of the event camera in its local frame $\{\mathcal{\Ebf}\}$ are defined as  ${}^\mathcal{E}\vbf_e\in \mathbb{R}^3$ and  ${}^\mathcal{E}\boldsymbol{\omega}_e \in \mathbb{R}^3$ respectively. 
Given a static landmark point in the camera frame ${}^\mathcal{E}\pbf_l\in \mathbb{R}^3$, the following equation holds:
\begin{equation}
\label{motion_eq}
 {}^\mathcal{E}\dot{\pbf_l} = \left[{}^\mathcal{E}\boldsymbol{\omega}_e\right]_{\times}{}^\mathcal{E}{\pbf_l} + {}^\mathcal{E}\vbf_e,
\end{equation}
where $\left[{}^\mathcal{E}\pmb\omega_e\right]_\times$ is the skew-symmetric matrix associated with angular velocity vector ${}^\mathcal{E}\boldsymbol{\omega}_e\in \mathbb{R}^3$, and ${}^\mathcal{E}\vbf_e(t)\in\mathbb{R}^3$ denotes the translational velocity.
We further define ${}^\mathcal{E}\pbf_l = \lambda{\xbf}$, where $\
\pmb{x}= \begin{bmatrix}
    \pmb{u}^\top &1
\end{bmatrix}^\top$ is the homogeneous coordinate form of $\pmb{u}$, then ${}^\mathcal{E}\dot{\pbf}_l = \dot{\lambda} {\xbf} + \lambda\dot{ {\xbf}}$. Substitute it into (\ref{motion_eq}), we have
\begin{equation}
		\label{motion_lambda}
 \dot{\xbf} = \left[{}^\mathcal{E}\boldsymbol{\omega}_e\right]_\times \xbf +\frac{1}{\lambda} {}^\mathcal{E}\vbf_e-\frac{\dot{\lambda}}{\lambda} {\boldsymbol{x}} .
\end{equation}
Now, multiplying both sides of (\ref{motion_lambda}) by $\left(\left[\xbf\right]_\times {}^\mathcal{E}\boldsymbol{v}_e\right)^\top$, we have
\begin{equation}
\label{velocity_flow_constraint}
    \left(\left[\xbf\right]_\times {}^\mathcal{E}\boldsymbol{v}_e\right)^\top\dot{\xbf} + \left(\left[\xbf\right]_\times {}^\mathcal{E}\boldsymbol{v}_e\right)^\top\left[\xbf\right]_\times {}^\mathcal{E}\boldsymbol{\omega}_e=0.
\end{equation}

The above equation defines the continuous epipolar constraint. Apparently, (\ref{velocity_flow_constraint}) does not depend on the 3D position of any world point, and only on the 2D observations of the world point. 
Under the assumption of brightness constancy, $	\dot \xbf(t) $ can be approximated by the optical flow $\dot{\ubf}(t)$ obtained from the event camera.  

The velocity of the camera in its own frame can be calculated from the radar velocity as 

\begin{equation}
 {}^\mathcal{E}\boldsymbol{v}_e = {}^{\mathcal{E}}\Rbf_{{r}}{}^{\mathcal{R}}\vbf_r + {}^{\mathcal{E}}\Rbf_{{r}}{}\left(^\mathcal{R}\pmb\omega_r\times {}^\mathcal{R}\lbf_{er}\right),
\end{equation}
where ${}^{\mathcal{E}}\Rbf_{{r}}$ is the rotation matrix from the radar frame to the event camera frame, respectively. ${}^\mathcal{R}\lbf_{er}$ denotes the displacement from radar to the camera. When ${}^\mathcal{R}\lbf_{er}$ is small enough, the velocity can be approximately obtained as ${}^\mathcal{E}\boldsymbol{v}_e = \Cbf_{\mathcal{R}}^{\mathcal{E}}{}^{\mathcal{R}}\vbf_r$.
where ${}^{\mathcal{R}}\vbf_r$ can be replaced with the closest radar ego-motion estimates in discrete time, assuming slow velocity variation.

Defining $\abf = \left(\left[\xbf\right]_\times \Cbf_{\mathcal{R}}^{\mathcal{E}}{}^{\mathcal{R}}\vbf_r\right)^\top \left[\xbf\right]_\times$, and $b = - \left(\left[\xbf(t)\right]_\times \Cbf_{\mathcal{R}}^{\mathcal{E}}{}^{\mathcal{R}}\vbf_r(t)\right)^\top \ubf(t)$ , equation (\ref{velocity_flow_constraint}) is simplified as 
\begin{equation}
	\abf {}^\mathcal{E}\pmb{\omega}_e(t) = \eta.
\end{equation}
Given $n\ge 3$ points satisfying constraint (\ref{velocity_flow_constraint}), and under the assumption that the angular velocity remains constant during a very short time period, we have the angular velocity equations as 
\begin{equation} 
	\label{angular_ls}
	 \Abf{}^\mathcal{E}\pmb{\omega}_e(t) = \pmb{\eta},
\end{equation}
where $\pmb{A}= \begin{bmatrix}
    \pmb{a}_1^\top &\pmb{a}_2^\top& \cdots &\pmb{a}_n  
\end{bmatrix}^\top$, $\pmb{b} = \begin{bmatrix}
    \eta_1& \eta_2& \cdots & \eta_n
\end{bmatrix}$
Then ${}^\mathcal{E}\pmb{\omega}_e(t)$ can be obtained in a least square fashion.

\section{Continuous-time Estimation}
In this section, a continuous-time ego-motion estimator is constructed to fuse the two types of instantaneous motion measurements, as shown in Fig.\ref{fig:event}.
We formulate the estimation problem as a nonlinear sliding-windowed estimation problem and solve it in a discrete-time manner.
\subsection{B-splines based trajectory}
To begin with, we represent the continuous-time velocities as cumulative B-splines. Specifically, we parameterize the translational and rotational velocities in two separate splines in the body local frame, which is represented by the continuous-time functions ${^\mathcal{B}}\vbf_b(t)\in\mathbb{R}^3$ and ${^\mathcal{B}}\pmb{\omega}_b(t)\in\mathbb{R}^3$. To simplify the problem, we assume that the body frame is aligned with the radar frame. The superscripts and subscripts are ignored, and we use $\vbf(t)$ and $\pmb{\omega}(t)$ for simplification. 

Consider the translational velocity function $\vbf(t)$ over a time period is of order $k$, and controlled by points $\vbf_i$, then it can be represented as 
\begin{equation}
		\label{trans_spline}
	\vbf(t)=\vbf_i+\sum_{j=1}^{k-1} \lambda^v_{j}(t) \cdot \Delta \vbf_{ij},
\end{equation}
where $ \lambda^v_j(t) $ is constant coefficient which depends on the order $k$, and $\Delta \vbf_{ij}\triangleq \vbf_{i+j} -\vbf_{i+j-1}\in \mathbb{R}^3$. 
Similarly, we define the rotational velocity as 
\begin{equation}
	\label{rot_spline}
\pmb{\omega}(t)=\pmb{\omega}_i +\sum_{j=1}^{k-1} \lambda^\omega_j(t) \cdot \Delta \pmb{\omega}_{ij}, 
\end{equation}
with $\Delta\pmb{\omega}_{ij}\triangleq \pmb{\omega}_{i+j} -\pmb{\omega}_{i+j-1}\in \mathbb{R}^3$.
\subsection{Sliding-windowed Optimization}
The objective of our estimator is to simultaneously minimize two factors: 1) The predicted translational velocities should be consistent with those measured from radar Doppler, and 2) the predicted angular velocity should be consistent with the optical flow obtained from the event camera based on Eq. (\ref{velocity_flow_constraint}).

To begin with, we define the sliding-window state to be estimated at time instance $k$ as follows:
\begin{equation}
    \mathcal{X}_k \triangleq \left\{\vbf^k_{j},\pmb{\omega}^k_{j}\right\}_{j=1:m}
\end{equation}
where $\vbf^k_{j},\pmb{\omega}^k_{j}$ denote the control points of $m-3$ B-splines at time instance $k$.
\begin{figure}
	\centering
	\includegraphics[width=0.95\linewidth]{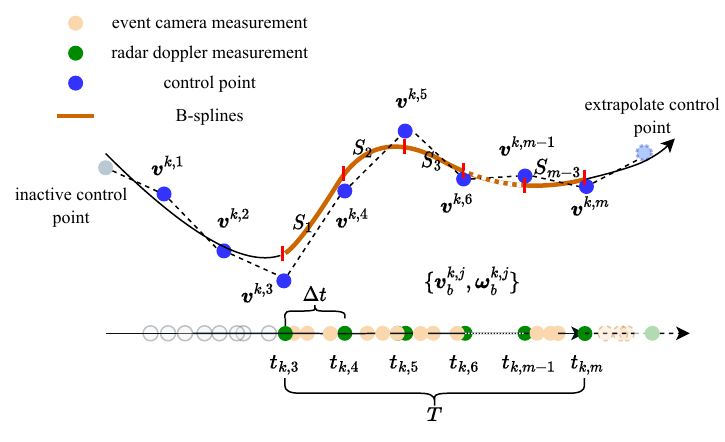}
	\caption{The B-splines control points distribution and measurements.}
	\label{fig:event}
    \vspace{-0.6 cm}
\end{figure}

Our sliding window-based optimization is illustrated in Fig.\ref{fig:event}. 
We consider using all measurements that fall into the period $\left[t_{k,3}, t_{k,m}\right]$. The objective function is formulated as
\begin{equation}
	\arg\min\limits_{\mathcal{X}^k} \alpha_{r}(\mathcal{X}^k) + \alpha_{e}(\mathcal{X}^k) + \alpha_{\text{prior}}(\mathcal{X}^k),
\end{equation}
where $\alpha_{r}(\cdot), \alpha_{e}(\cdot), \alpha_{\text{prior}}(\cdot)$ are the radar, event camera, and prior information terms, respectively.

\subsection{Measurement Models}
In this part, we formulate the measurement residuals related to the radar and the event camera. 
\subsubsection{Doppler Velocity Measurement}
We consider the radar measurement obtained from \ref{velocity_sec}, ${}^{\mathcal{R}}\hat{\vbf}_{\text{r}}$, which includes the true linear velocity corrupted by a constant bias and a time-varying noise, namely
\begin{equation}
{}^{\mathcal{R}}\hat{\vbf}_{r}^k = {}^{\mathcal{R}}\vbf^k_{r,true} + 
\boldsymbol{\xi}_v^k,
\end{equation} 
where 
$\boldsymbol{\xi}_v^k$ is a Gaussian white noise $\boldsymbol{\xi}_{\vbf}^k\sim \mathcal{N}(\mathbf{0},\Sigma_{\vbf} )$.
Specifically, in this part we directly integrate the velocity obtained from Doppler measurements in a loosely-coupled manner to simplify the problem. All points within a scan are used to first calculate the ego-motion velocity, then the velocity error term can be calculated as 
\begin{equation}
\rbf^k_{v} = {}^{\mathcal{R}}\hat{\vbf}_{r}^k - {^\mathcal{R}}\bar{\vbf}_r^k,
\end{equation}
where ${}^\mathcal{R}\bar{\vbf}^k_r$ is the interpolated/extrapolated pseudo measurement based on current estimates. 

Then the cost term w.r.t. radar $\alpha_r(\cdot)$ is defined as 
\begin{equation}
\alpha_r = \sum_{j\in M_r^k} \left\| \rbf_v^j \right\|_{\pmb\Sigma_v^{-1}}^2,
\end{equation} 
where $M_r^k$ is the set of active radar velocity measurements within the sliding window, and $\pmb\Sigma_v$ is the covariance matrix of the radar velocity measurement.


\subsubsection{Optical Flow Measurement}
In the loosely coupled fashion, an angular velocity measurement is first obtained based on Eq.\ref{angular_ls}. Without loss of generality, we consider the measurement to be corrupted by a noise term $\pmb{\xi}_{e}\sim \mathcal{N}(\mathbf{0}, \Sigma_{\pmb{\omega}})$. Then the angular velocity measurement can be modeled as
 \begin{equation}
	{}^{\mathcal{E}}\boldsymbol{\omega}^{k'}_{\text{e}} = {}^{\mathcal{E}}\boldsymbol{\omega}^{k'}_{\text{true}}  + \boldsymbol{\eta}_{\boldsymbol{\omega}}^{k'},
 \end{equation} 
 
The superscript $k'$ is used to denote the time instance when an angular velocity measurement is obtained. Note that we use a time instance variable $k'$ to denote the time instance for the event camera due to its asynchronous sensing mechanism.

We define the residual of the angular velocity as
\begin{equation}
    \rbf_e^{k'} = {}^{\mathcal{E}}\hat{\boldsymbol{\omega}}_{e} -{}^{\mathcal{E}}\bar{\boldsymbol{\omega}}_{e}.
\end{equation}
Then 
we can arrived at the cost function w.r.t. event camera $\alpha_{e}$ as
 \begin{equation}
	\alpha_{e} = \sum_{j\in M_{e}^k} \left\| \rbf_{e} \right\|_{\pmb\Sigma_{e}^{-1}}^2,
 \end{equation}
 where $M_e^k$ is the set of all active event measurements within the sliding window.

\subsubsection{Prior information}
Besides the above two factors, we also integrate prior information during each sliding-windowed optimization. Specifically, the prior factors constrain the common control points between two consecutive sliding windows. Additionally, we consider the biases and time offset as slow-varying parameters. Therefore, the prior constraints on the following states:
\begin{equation}
\mathcal{X}^k_{p} \triangleq \{ \pmb\Phi^{k-1} \cap \pmb\Phi^{k} 
\}.
\end{equation}
Then the prior factor residual can be defined as 
\begin{equation}
 \rbf^k_{{p}} = \Hbf_{{p}}\mathcal{X} - \mathcal{X}^k_{p},
\end{equation}
where $\Hbf_{p}$ is the matrix to select overlapped terms of current states w.r.t. prior information. 
Then the prior constraint term $\alpha_{prior}$ can be defined as 
\begin{equation}
    \alpha_{prior} = \|\rbf_{p}\|^2.
\end{equation}

\begin{figure}[t]
    \centering
    \includegraphics[width=1.0\linewidth]{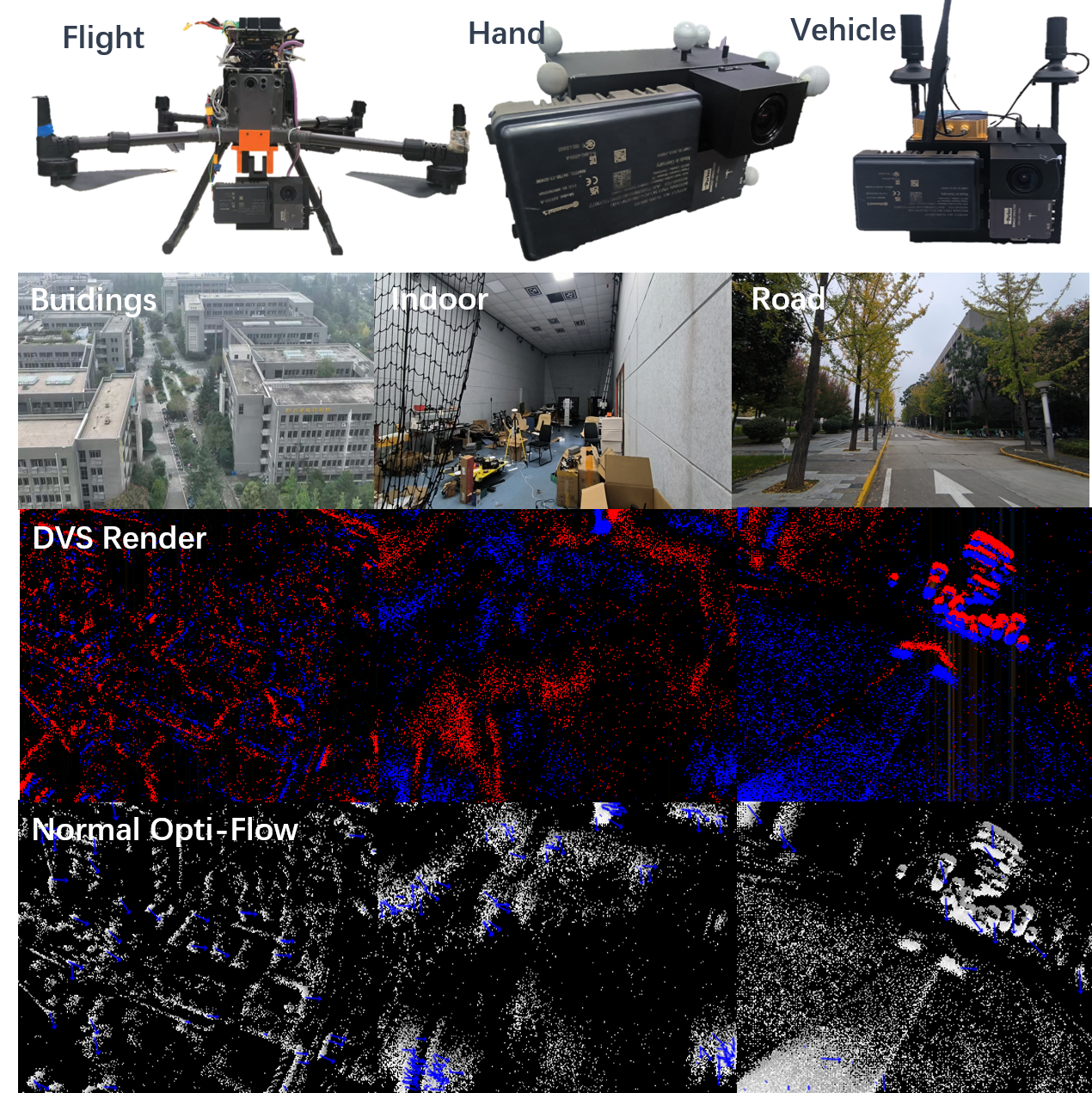}
    \caption{Experimental platform, scenery, dataset, and detection.}
    \label{fig:Platform}
    \vspace{-0.5cm}
\end{figure}

\section{Experiment}
\subsection{Platforms \& Datasets}
\label{sec:platform_setups}





To validate the proposed method, we integrate multiple sensors, including a DAVIS346 event camera, an ARS548 4D millimeter-wave radar, a Parker IMU, and an Intel NUC13, forming the core sensing module. Three experiments are conducted:
\begin{itemize}[leftmargin=*]
\item {Flight} The sensor suite is mounted on a DJI M300 for real-flight tests in semi-urban areas. Low-altitude, forward-view trajectories navigate through building clusters and narrow passages, to evaluate robustness under complex structural conditions. Ground truth is provided by RTK with a high-precision positioning network.

\item {Handheld} A handheld setup captures indoor data under aggressive motion and HDR lighting. The cluttered environments, featuring corridors and reflective surfaces, serve to evaluate the robustness of perception and motion estimation. Ground truth is obtained from the OptiTrack system.

\item{Vehicle} An electric vehicle conducts high-speed road experiments under day and night HDR conditions. Dynamic traffic scenes with moving vehicles, pedestrians, and illumination changes are used to assess system robustness and real-time performance. Ground truth is provided by RTK integrated with the positioning network.
\end{itemize}









We collected a total of 9 sequences and performed a statistical analysis of the ground-truth velocities. Table\ref{tab:twist_hdr_statistics} summarizes the linear and angular velocity statistics for the nine recorded sequences.

\begin{table}[h]
\centering
\caption{Twist and HDR Statistics for Sequences.}
\label{tab:twist_hdr_statistics}
\begin{tabular}{lccccc}
\toprule
File & Lin.Max & Lin.Avg & Ang.Max & Ang.Avg & HDR \\
\midrule
dji1   & 4.49 & 4.01 & 0.34 & 0.16 & 62 \\
dji2   & 4.51 & 3.47 & 0.54 & 0.16 & 63 \\
dji3   & 6.30 & 3.74 & 0.56 & 0.18 & 62 \\
hand1  & 4.20 & 0.92 & 0.42 & 0.16 & 89 \\
hand2  & 2.32 & 0.87 & 0.45 & 0.17 & 92 \\
hand3  & 4.39 & 1.17 & 0.41 & 0.16 & 97 \\
road1  & 7.64 & \textbf{6.32} & \underline{0.55} & \textbf{0.28} & 96 \\
road2  & \textbf{7.75} & \underline{5.83} & 0.54 & \underline{0.27} & \underline{101} \\
road3  & \underline{7.51} & 5.40 & \textbf{0.65} & 0.17 & \textbf{109} \\
\bottomrule
\end{tabular}
\vspace{1pt}
{
\begin{flushleft}
\footnotesize Notation: Linear velocity in m/s, angular velocity in rad/s, HDR in dB.\\
\textbf{Bold} indicates the best value \\
\underline{Underline} indicates the second-best value.
\end{flushleft}
\vspace{-0.2cm}
}
\end{table}

For the dji flight sequences, linear velocities range from 3.47 to 4.01 m/s and angular velocities from 0.16 to 0.56 rad/s, reflecting moderate low-altitude flights over building clusters and narrow passages with a top-down view. These trajectories provide structural and visual complexity for validating the method under real-flight conditions. The HDR in these sequences ranges around 62 to 63 dB, representing typical outdoor lighting conditions for standard event camera perception.

Handheld sequences exhibit lower motion, with linear velocities below 1.2 m/s and angular velocities around 0.16 to 0.17 rad/s, corresponding to aggressive indoor hand motion within cluttered scenes. The HDR is estimated around 89 to 97 dB, indicating darker indoor lighting and emphasizing robustness to rapid viewpoint changes and low-light conditions.

Road sequences, collected using an electric vehicle, reach up to 7.75 m/s and 0.65 rad/s, involving high-speed motion in dynamic traffic and varying illumination, designed to test system stability and real-time performance under fast outdoor conditions. HDR in these sequences is higher, around 96 to 109 dB, reflecting strong sunlight and overexposed regions, which challenge event-based perception under high dynamic range scenarios.

\begin{table}[t]
\centering
\caption{Characteristic of Algorithms}
\label{tab:algorithms_characteristic}
\begin{tabular}{>{\centering\arraybackslash}m{3.5cm}ccccccc}
\toprule
\multirow{2}{*}{Algorithm} & \multicolumn{4}{c}{Input} & \multicolumn{3}{c}{Output} \\
\cmidrule(lr){2-5} \cmidrule(lr){6-8}
 & E & R & V & I & L & A & P \\
\midrule
CMAX\cite{guo2024cmax}       & \checkmark &  &  &  &  & \checkmark &  \\
RIO\cite{10833750}        &  & \checkmark &  & \checkmark & \checkmark &  & \checkmark \\
River\cite{chen2024river}      &  & \checkmark &  & \checkmark & \checkmark &  & \checkmark \\
PLEVIO\cite{guan2023pl}     & \checkmark & & \checkmark & \checkmark & \checkmark &  & \checkmark \\
DEVO\cite{klenk2023devo}       & \checkmark &  &  &  &  &  & \checkmark \\
TE(front, \textbf{proposed})  &  & \checkmark & & \checkmark & \checkmark & \checkmark &  \\
TE(back, \textbf{proposed})   &  & \checkmark & & \checkmark & \checkmark & \checkmark &  \\
\bottomrule
\end{tabular}

\vspace{4pt} 
{\raggedright\footnotesize 
Notation: E: Event; R: Radar; V: Camera; I: IMU; L: Linear velocity; A: Angular velocity; P: Pose.\par}
\vspace{-0.8cm}
\end{table}

\begin{table}[b]
\centering
\caption{\textsc{Linear Velocity Error} (\textnormal{m/s})}
\label{tab:linear_velocity_error}
\resizebox{0.485\textwidth}{!}{
\begin{tabular}{>{\raggedright\arraybackslash}p{1.5cm} c c c c c c}
\hline
\rotatebox{45}{Seq} & \rotatebox{45}{RIO} & \rotatebox{45}{River} & \rotatebox{45}{PLEVIO} & \rotatebox{45}{DEVO} & \rotatebox{45}{TE(front)} & \rotatebox{45}{TE(back)} \\ \hline
dji1 & 1.22 & 1.99 & $\times$ & 1.07 & \underline{0.27} & \textbf{0.22} \\
dji2 & 3.18 & 2.77 & $\times$  & 1.77 & \underline{0.18} & \textbf{0.16} \\
dji3 & $\times$ & 1.29 & $\times$  & 0.98 & \underline{0.09} & \textbf{0.05} \\
hand1 & $\times$ & $\times$ & $\times$  & $\times$ & \underline{0.67} & \textbf{0.40} \\
hand2 & $\times$ & 2.91 & $\times$ & 1.38 & \underline{0.10} & \textbf{0.06} \\
hand3 & $\times$ & 1.63 & $\times$ & 0.88 & \underline{0.75} & \textbf{0.44} \\
road1 & $\times$ & $\times$ & $\times$ & $\times$ & \underline{0.17} & \textbf{0.11} \\
road2 & $\times$ & $\times$ & 0.40  & 1.46 & \underline{0.17} & \textbf{0.12} \\
road3 & $\times$ & 2.83 & 0.24  & $\times$ & \underline{0.14} & \textbf{0.11} \\ \hline
\end{tabular}
}
{
\begin{flushleft}
\footnotesize Notation: \textbf{Bold} indicates the best value \\
\underline{Underline} indicates the second-best value.
\end{flushleft}
}
\end{table}

\subsection{Algorithms}
We selected several representative algorithms for comparison, covering state-of-the-art methods across different sensor modalities. CMAX-SLAM (CMAX) \cite{guo2024cmax} focuses on angular velocity estimation, RIO\cite{10833750} is a radar–inertial odometry approach, River\cite{chen2024river} estimates linear velocity using radar–inertial fusion, PL-EVIO (PLEVIO)\cite{guan2023pl} is an event-based visual–inertial odometry method, and DEVO\cite{klenk2023devo} is a state-of-the-art event-only odometry approach. Collectively, these methods represent representative sensor and algorithmic combinations for high-dynamic pose and velocity estimation. Combined with our proposed Twist-Estimator (TE), including front-end (front) and back-end (back) results, we evaluate all methods on our self-recorded datasets. Input and primary output types are summarized in Table~\ref{tab:algorithms_characteristic}.
Compared to adaptive tuning methods such as NeuroMHE \cite{wang2023neural}, we set fixed residual weights for linear and angular velocities to 400 and 800 for system simplification. The relatively large weights help prolong the optimization process and improve the resolution of small residuals. For the spline representation, we use cubic B-splines ($k=3$), the control points in this formulation are fully observable in the velocity state, making the spline representation effectively as observable as sensors measurements.

\begin{table}[b]
\centering
\caption{\textsc{Angular Velocity Error} (\textnormal{rad/s})}
\label{tab:angular_velocity_error}
\resizebox{0.5\textwidth}{!}{
\begin{tabular}{cccccccc}
\hline
\rotatebox{45}{Seq} & \rotatebox{45}{CMAX} & \rotatebox{45}{RIO} & \rotatebox{45}{River} & \rotatebox{45}{PLEVIO} & \rotatebox{45}{DEVO} & \rotatebox{45}{TE(front)} & \rotatebox{45}{TE(back)} \\ \hline
dji1 & $\times$ & 0.14 & 0.09 & $\times$  & 0.21 &  \underline{0.07} & \textbf{0.06} \\
dji2 & \underline{0.08} & 0.10 & 0.18 & $\times$  & 0.34 & 0.10 & \textbf{0.07} \\
dji3 & 0.07 & 0.08 & 0.11 & $\times$  & 0.14 & \underline{0.05} & \textbf{0.03} \\
hand1 & 0.17 & $\times$ & 0.15 & $\times$  & $\times$ & \underline{0.11} & \textbf{0.08} \\
hand2 & 0.15 & 0.10 & 0.15 & $\times$  & 0.16 & \underline{0.04} & \textbf{0.01} \\
hand3 & 0.17 & 0.15 & 0.15 & $\times$ & 0.26 & \underline{0.11} & \textbf{0.06} \\
road1 & 0.18 & $\times$ & 0.17 & \textbf{0.16}  & $\times$ & 0.19 & 0.17 \\
road2 & 0.17 & $\times$ & 0.17 & 0.17 & \underline{0.16} & 0.17 & \textbf{0.15} \\
road3 & 0.12 & 0.09 & 0.12 & 0.11  & $\times$ & 0.09 & \textbf{0.08} \\ \hline
\end{tabular}
}
{
\begin{flushleft}
\footnotesize Notation: \textbf{Bold} indicates the best value \\
\underline{Underline} indicates the second-best value.
\end{flushleft}
}
\end{table}

Velocity estimation is measured by the Absolute Velocity Error (AVE), capturing both magnitude and temporal trends. Let $\mathbf{x}_{\text{est}}$ and $\mathbf{x}_{\text{gt}}$ denote the estimated and ground-truth velocities (linear or angular), with instantaneous error
\[\text{AVE} = \frac{1}{N} \sum_{i=1}^{N} \left\| \mathbf{x}_{\text{est},i} - \mathbf{x}_{\text{gt},i} \right\|_2,\]
where $N$ denotes the length of the evaluated sequence.

\vspace{-0.2cm}

\subsection{Linear Velocity Evaluation}
Table~\ref{tab:linear_velocity_error} summarizes linear velocity estimation across all sequences, and illustrative examples of the error curves based on dataset $\mathtt{hand2}$ and $\mathtt{road2}$ are provided in Fig. \ref{fig:hand2_linear} and \ref{fig:road2_linear}, respectively.
\begin{itemize}[leftmargin=*]

\begin{figure}[t]
  \centering
  \begin{subfigure}[b]{0.47\textwidth}
    \includegraphics[width=\linewidth]{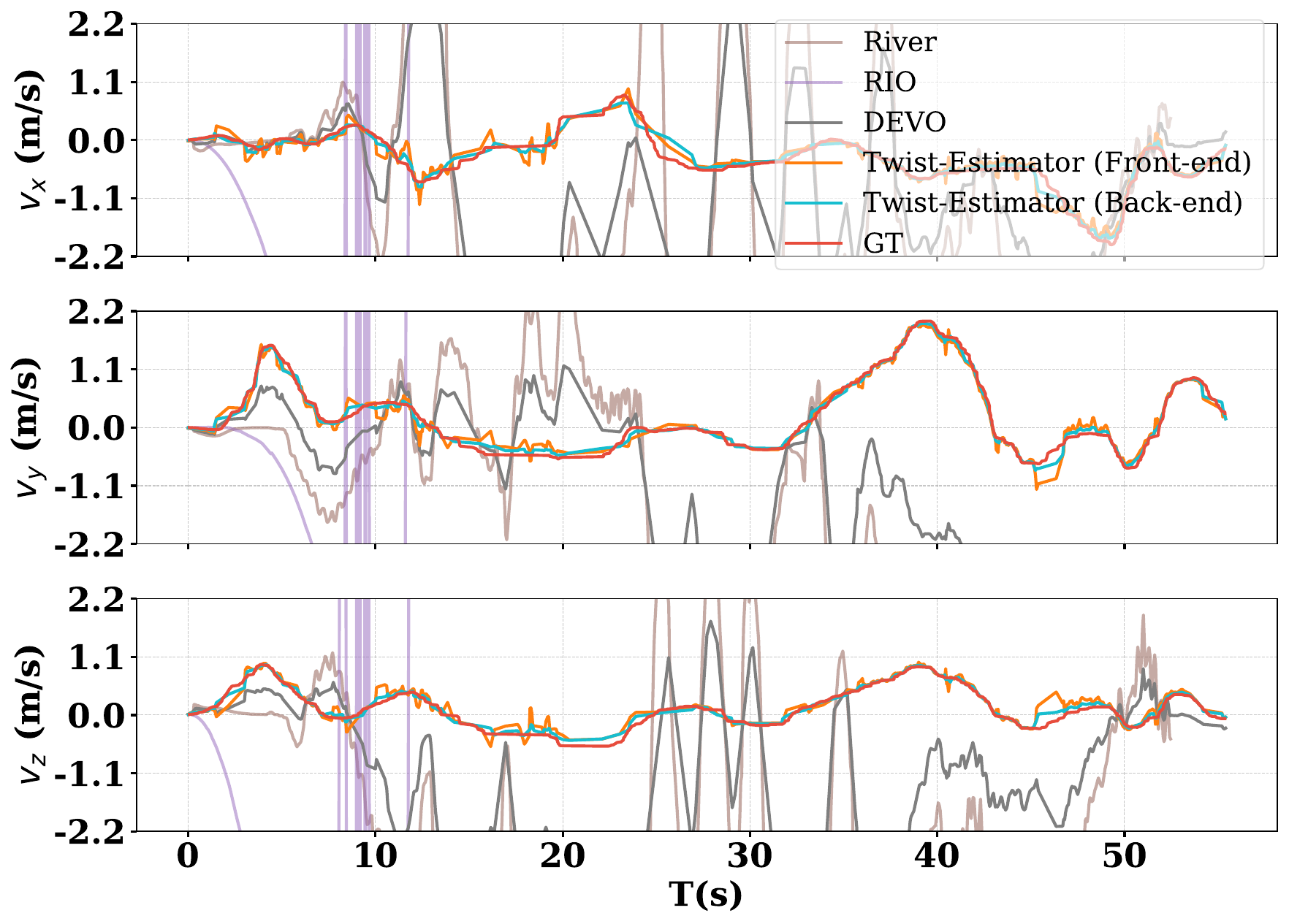}
    \caption{$\mathtt{hand2}$ Linear velocity.}
    \label{fig:hand2_linear}
  \end{subfigure}
  \hfill
  \begin{subfigure}[b]{0.47\textwidth}
    \includegraphics[width=\linewidth]{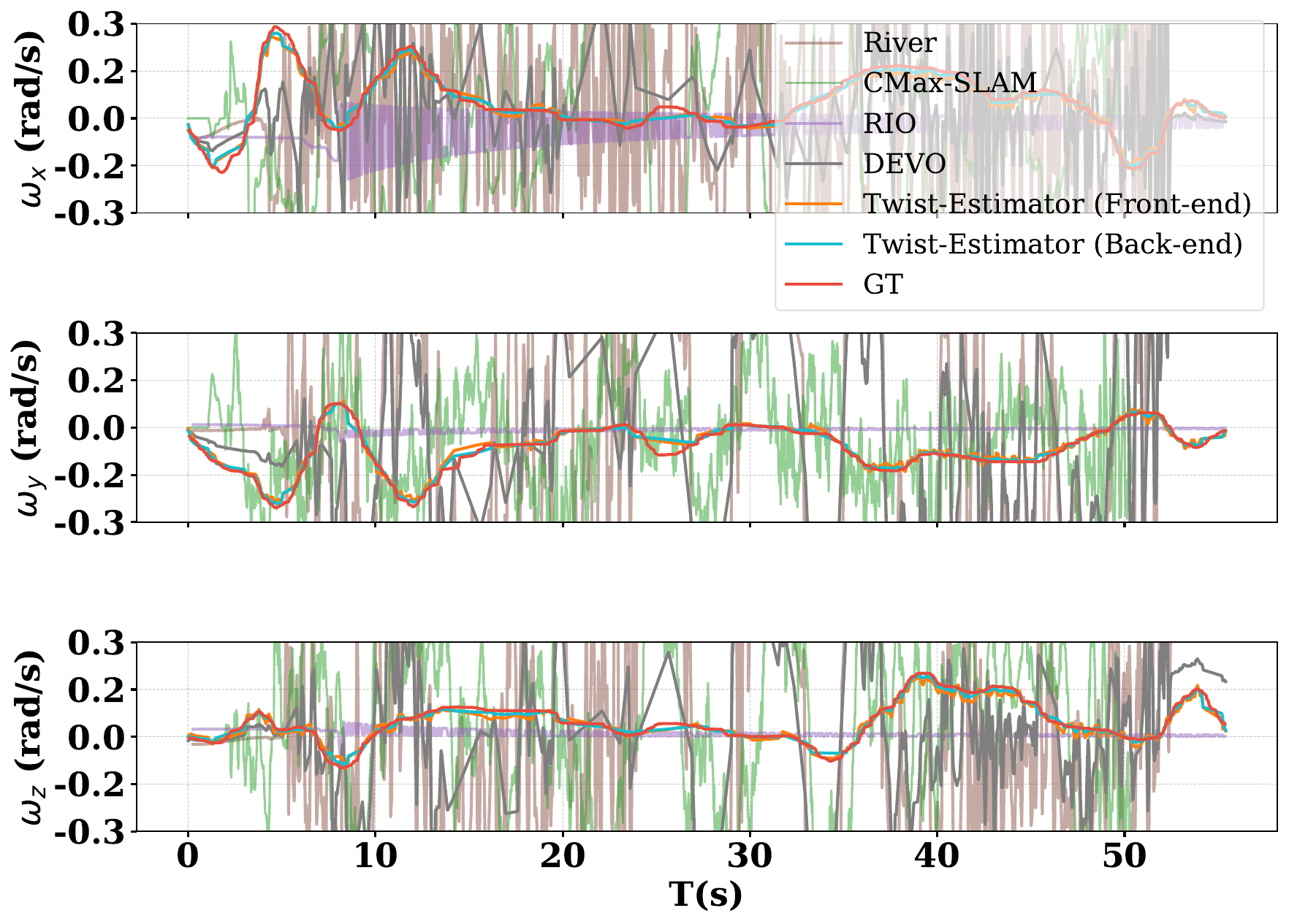}
    \caption{$\mathtt{hand2}$ Angular velocity.}
    \label{fig:hand2_angular}
  \end{subfigure}
  \caption{Twist Evaluation of Algorithms on Seq. $\mathtt{hand2}$.
  }
  \label{fig:velocity_comparison}
  \vspace{-0.6 cm}
\end{figure}

\begin{figure}[t]
  \centering
  \begin{subfigure}[b]{0.47\textwidth}
    \includegraphics[width=\linewidth]{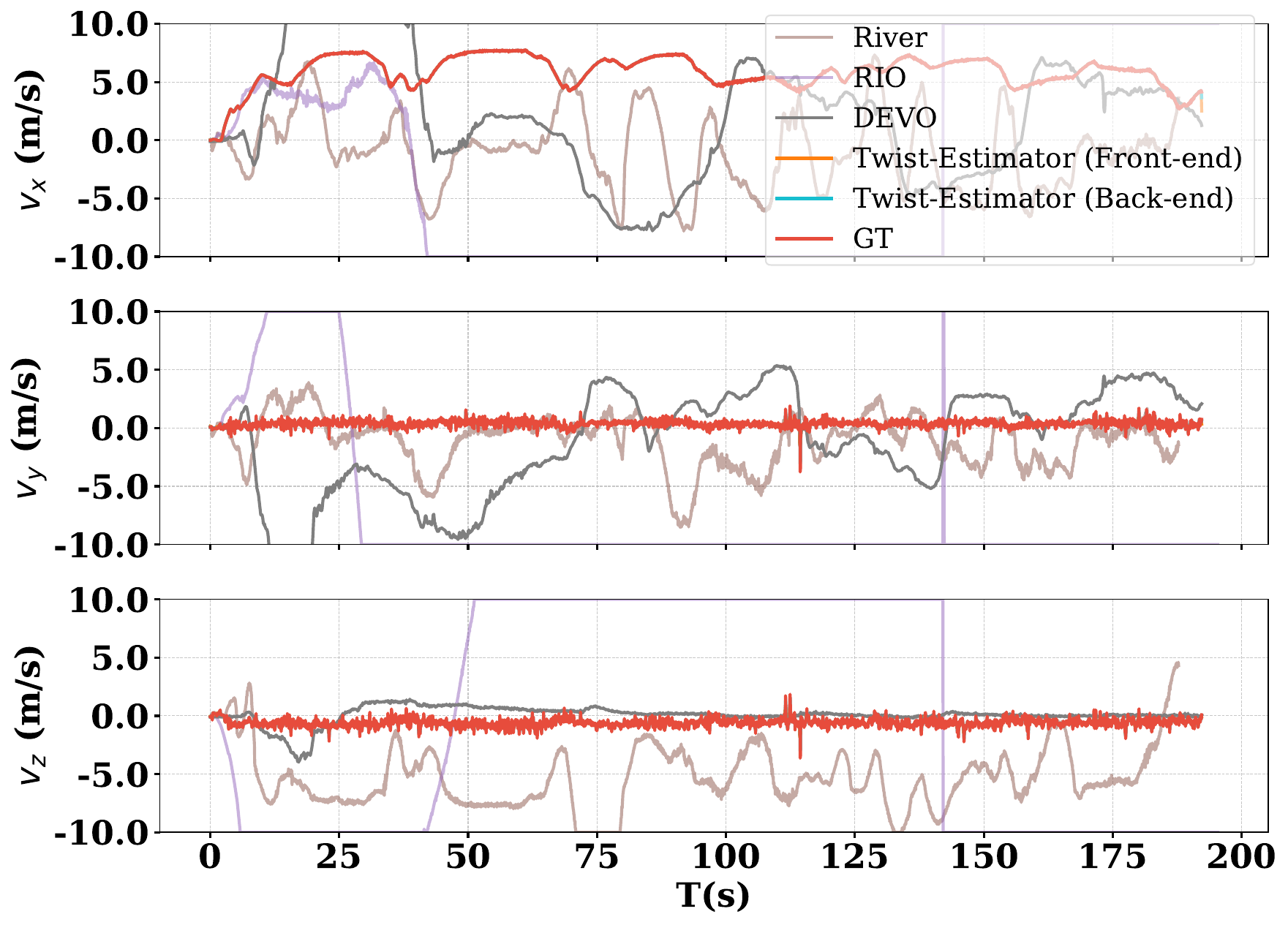}
    \caption{$\mathtt{road2}$ Linear velocity.}
    \label{fig:road2_linear}
  \end{subfigure}
  \hfill
  \begin{subfigure}[b]{0.47\textwidth}
    \includegraphics[width=\linewidth]{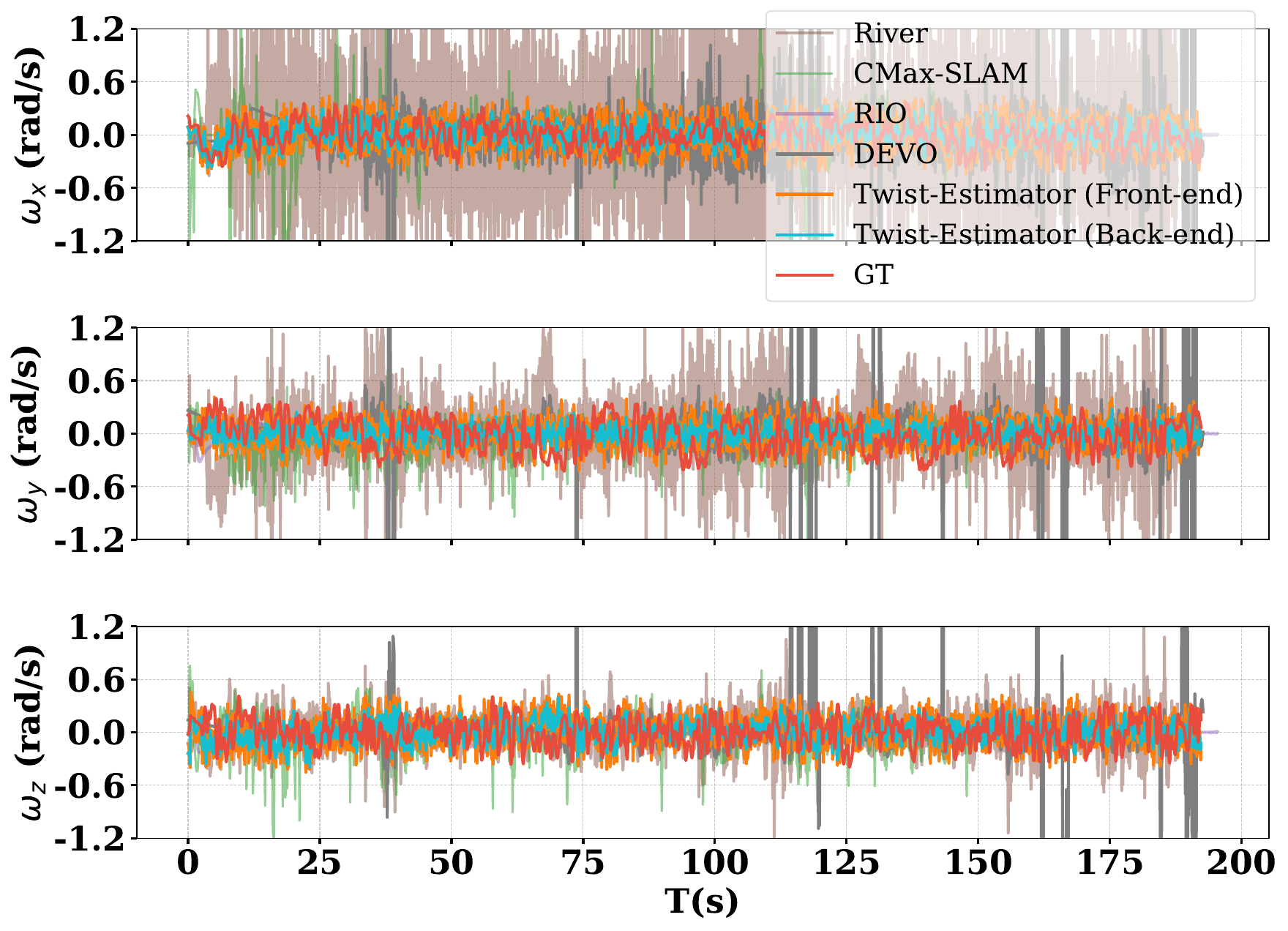}
    \caption{$\mathtt{road2}$ Angular velocity.}
    \label{fig:road2_angular}
  \end{subfigure}
  \caption{Twist Evaluation of Algorithms on Seq. $\mathtt{road2}$.
  }
  \label{fig:velocity_comparison_dji3}
  \vspace{-0.5 cm}
\end{figure}

\item {RIO} relies on IMU acceleration integration and radar scan matching. It exhibits large errors in most sequences due to drift and unreliable radar constraints, achieving moderate accuracy only in structured scenes like $\mathtt{dji1}$ and $\mathtt{dji2}$. In unobvious structure or water surface reflective scenes such as $\mathtt{dji3}$, radar matching is unreliable, causing unstable estimates.

\item {River} fuses radar Doppler with IMU for direct velocity estimation. Inconsistencies between Doppler and inertial measurements reduce fusion effectiveness, and performance degrades under significant IMU bias or vibrations, especially in high-speed road sequences.

\item {PLEVIO} uses event-based point–line feature tracking initialized by intensity gradients. Sparse, low-contrast, or spatially sparse events in handheld and aerial sequences often cause tracking failures. Indoor clutter and transient corridor transitions further limit reliable estimation.

\item{DEVO} is an event-only approach, relies on reconstructed event patches. While avoiding IMU drift allows better performance in limited-motion indoor scenes, accuracy degrades in aerial or road sequences with larger motion ranges or depth variation, where the pre-trained model fails to generalize.

\item{Proposed TE} estimates velocity directly from external observations and avoids IMU integration. TE(front) uses radar Doppler for high accuracy, and the addition of event optical flow constrains linear motion, which reduces drift. Compared to PLEVIO, our method computes optical flow over small $3\times3$ neighborhoods, providing robust constraints even under sparse or low-contrast events. TE(back) incorporates marginalization-based smoothing to filter spikes, and yields conservative yet stable estimates. As a result, TE(back) achieves the lowest errors in sequences such as $\mathtt{dji1,2,3}$ and $\mathtt{road1,2,3}$, consistently outperforming existing baselines across aerial, handheld, and vehicle platforms under moderate and aggressive motion.
\end{itemize}

\vspace{-0.5 cm}
\subsection{Angular Velocity Evaluation}
Table~\ref{tab:angular_velocity_error} presents angular velocity estimation results across all sequences, and illustrative examples of the error curves based on dataset $\mathtt{hand2}$ and $\mathtt{road2}$ are provided in Fig. \ref{fig:hand2_angular} and \ref{fig:road2_angular}, respectively.
\begin{itemize}[leftmargin=*]
    \item RIO and River rely on IMU high-frequency angular velocity measurements, providing generally accurate references across sequences. Their performance benefits from inertial sensing, particularly in structured aerial or road scenes, but can degrade under high vibrations or inconsistent radar constraints.
    \item CMAX depends primarily on structured event-based features for rotational estimation. It performs well in aerial sequences such as $\mathtt{dji2}$ and $\mathtt{dji3}$, where events exhibit clear structure and flight is dominated by near-pure rotational motion. In $\mathtt{dji1}$, the low flight speed and angular motion reduce the effectiveness of CMAX. Indoor sequences with cluttered, spatially dispersed edges, or road sequences with large depth variations, violate the pure-rotation assumption, leading to larger errors.
    \item PLEVIO also depends on event-based point–line feature tracking. While event structures are relatively clear in aerial sequences, sparse or unstable features limit reliable angular estimation. Handheld and indoor sequences with low-contrast or dispersed events further reduce tracking accuracy.
    \item DEVO, an event-only method, avoids IMU-induced drift and performs well in indoor sequences with limited motion. However, in aerial and high-speed road sequences, large motion ranges, depth changes, and mismatches with training data lead to degraded performance.
    \item Proposed TE estimates angular velocity directly from external observations, avoiding IMU integration. TE(front) achieves near-optimal performance in most sequences, and TE(back), with marginalization-based smoothing, consistently attains the lowest errors. Across aerial, handheld, and vehicle sequences, TE surpasses the inertial angular velocity reference in many cases, though some high-vibration scenes (e.g., $\mathtt{dji2}$, $\mathtt{road1,2,3}$) are affected by environmental disturbances. Overall, TE demonstrates robust, accurate angular estimation across diverse platforms and motion dynamics.
\end{itemize}

\vspace{-0.3 cm}
\subsection{Supplemental trajectory evaluation}
\begin{figure}[t]
    \centering
    \includegraphics[width=0.5\textwidth]{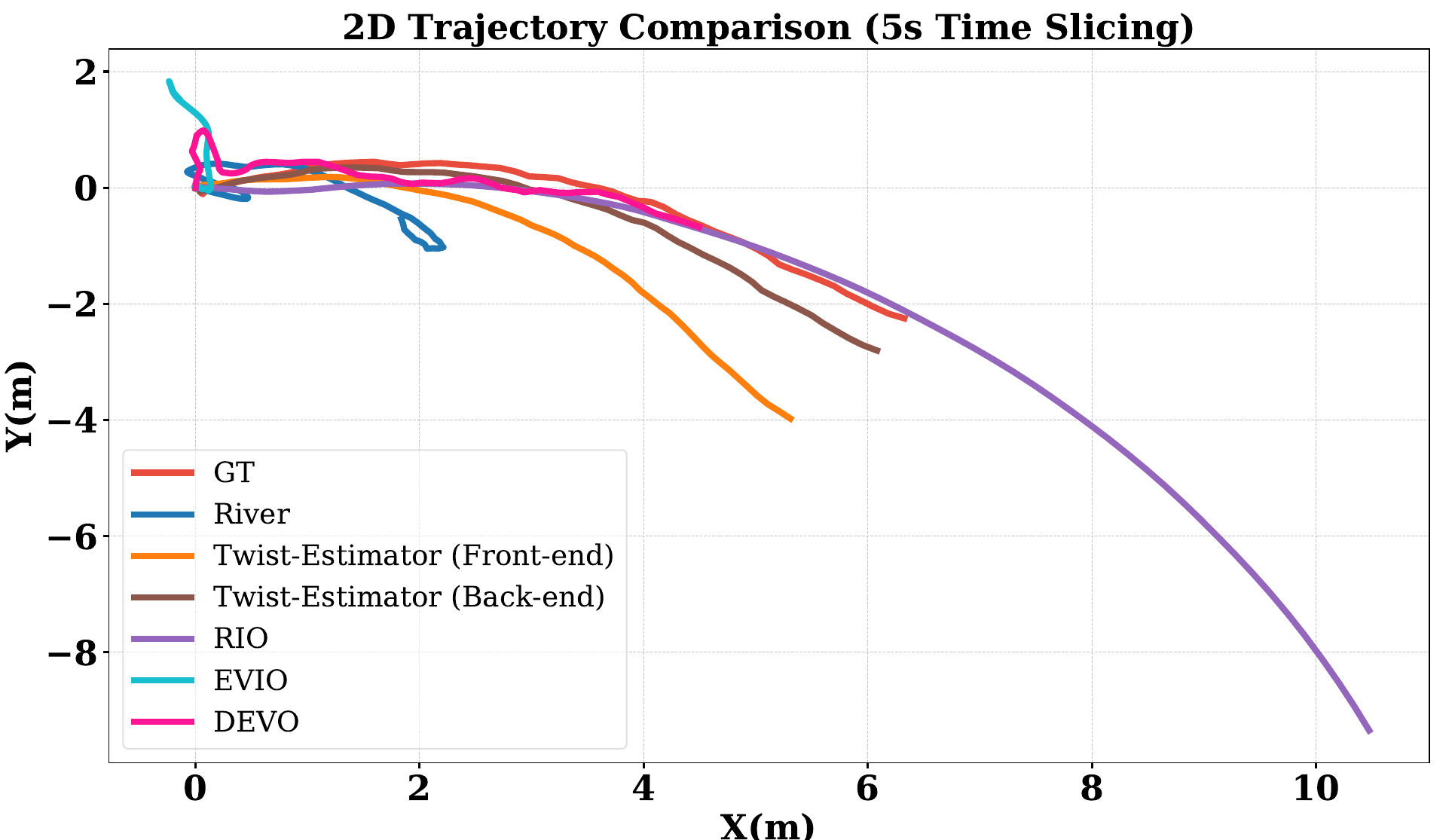}
    \caption{Velocity-Integrated Trajectory Comparison over 5 s in $\mathtt{road2}$}
    \label{fig:2d_traj}
    \vspace{-0.6 cm}
\end{figure}

Another promising application of instantaneous velocity estimation is pose estimation. Integrating instantaneous velocities over time mitigates the accumulation of dynamic errors in higher-dimensional motion, making it an effective approach for short-term localization. To evaluate this capability, we conducted an extended experiment in which poses obtained by integrating the velocity outputs of our proposed method over a 5-second interval were compared against the corresponding poses from baseline algorithms, with all trajectories first aligned via SE(3) transformation, as illustrated in the figure \ref{fig:2d_traj}.

The experimental results demonstrate that the proposed method, particularly when the velocity estimates are optimized before integration, produces poses that most closely align with the ground truth. This indicates the significant potential of our approach for pose estimation, highlighting its suitability for highly dynamic scenarios where short-term accurate localization is critical.

\vspace{-0.5cm}
\section{Conclusion}
We propose a robust and efficient ego-motion estimation framework for dynamic, visually challenging environments. It avoids inertial data and complex feature matching, offering a lightweight solution for agile robots. Experiments show stable motion estimates where traditional methods struggle. Designed as a minimal 6-DoF velocity estimator, it can serve as an alternative to IMU-based approaches. {In future work, we plan to integrate IMU measurements to provide system redudency to more challenging environments.}

\appendices
\ifCLASSOPTIONcaptionsoff
  \newpage
\fi



\bibliographystyle{IEEEtran}
\bibliography{bibtex/bib/IEEEexample}
\end{document}